\pdfoutput=1

\documentclass[11pt]{article}

\usepackage{dezzai}

\usepackage{times}
\usepackage{latexsym}
\usepackage[T1]{fontenc}

\usepackage[utf8]{inputenc}

\usepackage{microtype}

\usepackage{inconsolata}

\usepackage{svg}
\graphicspath{ {./images/} }
\usepackage[title]{appendix}
\usepackage{xurl}

\usepackage{booktabs}  
\usepackage{wasysym}  
\usepackage{multirow}  
\usepackage[group-digits=integer,group-minimum-digits=4,group-separator={,},output-decimal-marker={.}]{siunitx}  

\title{Spanish Built Factual Freectianary (Spanish-BFF): the first AI-generated free dictionary}

\author{
Miguel Ortega-Martín\\
  dezzai, UCM \\
  \small\texttt{m.ortega@dezzai.com} \\
  \small\texttt{m.ortega@ucm.es} \\
  \And
  Óscar García-Sierra \\
  dezzai, UCM \\
  \small\texttt{oscar.garcia@dezzai.com} \\
  \small\texttt{oscarg02@ucm.es} \\
  \And
  Alfonso Ardoiz \\
  dezzai \\
  \small\texttt{alfonso.ardoiz@dezzai.com} \\
  \AND
  \textbf{Juan Carlos Armenteros} \\
  dezzai \\
  \small\texttt{juancarlos.armenteros@dezzai.com} \\
  \And
   \textbf{Jorge Álvarez} \\
  dezzai \\
  \small\texttt{jorge.alvarez@dezzai.com} \\
  \small\texttt{} \\
  \And 
  \textbf{Adrián Alonso} \\
  dezzai, URJC, DSL URJC\\
  \small\texttt{a.alonso@dezzai.com} \\
  \small\texttt{adrian.barriuso@urjc.es} \\
  }

\begin{document}

\maketitle

\begin{abstract}

Dictionaries are one of the oldest and most used linguistic resources. Building them is a complex task that, to the best of our knowledge, has yet to be explored with generative Large Language Models (LLMs). We introduce the "Spanish Built Factual Freectianary" (Spanish-BFF) as the first Spanish AI-generated dictionary. This first-of-its-kind free dictionary uses GPT-3. We also define future steps we aim to follow to improve this initial commitment to the field, such as more additional languages.

\end{abstract}

\section{Introduction}

Dictionaries are among the earliest and most extensively utilized linguistic resources. A dictionary is a list of alphabetically ordered words, although sorting is not always necessary. There are many types of dictionaries, such as monolingual, bilingual, general or domain-specific. Each kind has its peculiarities (for example, structure or content). Here we build a general-purpose monolingual Spanish dictionary, which provides the meaning of Spanish lemmas. While it tries to cover the totality of the lemmas utilized in this language, it may require more completeness. Nevertheless, as a free and open-source tool, we promote contributions to this initiative.

This work contains a broad review of lexicography (section \ref{sec:lexicography}); an introduction to Large Language Models (LLMs) in general and GPT-3 particularly (section \ref{sec:gpt3}); our experimental set-up (section \ref{sec:experimental-setup}); an analysis of the generated dictionary (section \ref{sec:results}); the future lines of this project (section \ref{sec:future-lines}); the conclusions (section \ref{sec:conclusions}); the limitations of our work (section \ref{sec:limitations}); and the Ethics Statement (section \ref{sec:ethics}). 

Our contributions to this paper are the following:

\begin{itemize}
    \item We build, to the best of our knowledge, the first free AI-generated dictionary. In particular, it contains Spanish lemmas and definitions generated with GPT-3. 
    \item We establish the future lines of work for the field and are also pleased to provide an effective procedure to achieve further goals.
\end{itemize}

\section{Lexicography\label{sec:lexicography}}

Lexicography is the study of dictionaries and how they are compiled. We can distinguish between theoretical lexicography, which includes theories about the structure and contents of dictionaries, and practical lexicography, which deals with creating concrete dictionaries \cite{bergenholtz12}. 

Building a dictionary is an arduous task that follows some guidelines and lexicographic principles, ensuring an efficient consultation and understanding \cite{jackson13}. A dictionary's structure has three major components: outside matter (as additional resources or use guidelines), macro-structure, and micro-structure \cite{hausmann89, kirkness04}. Macro-structure refers to the list of lemmas and their organization. The size and structure of this list depend on the type and field of dictionary \cite{hausmann89, kirkness04}. Micro-structure refers to the linguistic data that each entry contains.

Definitions are the essential elements of dictionaries. Traditionally a good definition of a lemma contains the following information in order of appearance: generic term, Part of Speech (POS) tags (the class the term belongs to), a list of senses according to a predefined sorting rule and eventually, some use cases. Furthermore, we can find other linguistic notes, like spelling and pronunciation, base and inflected forms, morphological information and other semantic knowledge (like synonyms, antonyms, hypernyms, or hyponyms) \cite{kirkness04}.

As in many other linguistic fields, computers can contribute to lexicography, where "the electronic storage of vast textual material in corpora and the varied electronic presentation of lexicological and lexicographical work represent a quantum leap" \cite{kirkness04}. Computational lexicography has focused on human-annotated dictionaries and lexicons generation from large amounts of raw text. Traditionally, electronic dictionaries require considerable human, economic and computational costs. As systems relied on computational lexicons to match the tokens from an input text, dictionaries significantly contributed to NLP. But modern neural LMs work without them and build tailored vocabularies optimizing computational costs instead. Consequently, LMs facilitate an essential open resource to the broad public where definitions stop enduring to evolve according to trending use. Regardless, this flexibility assumes the cost of generalizing.

Recently, some approaches employed dictionaries to create word embeddings and benefit from lexicographic information that typically is not utilized when training models on Internet corpora. Following this line, \cite{hill16} and \cite{ortega21} use the ensemble of words from the definition to compute its embedding. Besides, Definition Modelling (DM) is an NLP task that proposes generating meanings from word embeddings \cite{noraset17}. It uses Recurrent Neural Networks (RNNs) \cite{schmidt19} and performs qualitative and quantitative error analysis of the generated definition, and \cite{bevilacqua20} develops contextual glosses from words and phrases using a BART model \cite{lewis19}.

As far as we are concerned, the application of generative LMs to construct comprehensive dictionaries has yet to be investigated.

\section{GPT-3\label{sec:gpt3}}

Large Language Models (LLMs) are at the forefront of NLP. Particularly, GPT-3 \cite{brown20} is one of the most famous encoder-decoder models. It generates text from a given input based on vector representations of words or parts of words. Most recent versions of this model aim to employ user intents to boost their performance. For instance, IntructGPT \cite{ouyang22} is a fine-tuned model of GPT-3 with supervised learning, \num{1.3} billion parameters and human feedback. ChatGPT\footnote{\url{https://openai.com/blog/chatgpt/}} is another fine-tuned version of GPT-3 having \num{175} million parameters, yet trained to interact with users. Prompting \cite{ouyang22} is crucial in all of them. Indeed, a prompt is a piece of text used to reduce the context of the input to improve the quality of the generated text.  

Although generative models have been widely used for many NLP tasks, their capabilities for assembling an entire dictionary still need to be investigated. For example, \cite{malkin21} explore GPT-3 to provide a meaning for new words.

\section{Experimental set-up\label{sec:experimental-setup}}

We use a curated list of \num{66353} unique forged Spanish lemmas (including neologisms) and generate a single definition for each. To benchmark the results,  we parse the output of queries of these lemmas to the "Diccionario de la Lengua Española" \footnote{\url{https://dle.rae.es/}} (DLE), which aims to contain all Spanish words. We do not store, manipulate, or intend to make any commercial use of these outputs whatsoever. This procedure is exclusively for research purposes: to contrast the performance of the dictionary proposed against a trusted source. Our first proposal neglects homonymy and polysemy. Although POS tags of the lemmas are currently ignored, we restrict them to nouns, verbs, adjectives and adverbs. However, we will aim to use this information for further improvements.

\begin{table*}[t]
\centering
\begin{tabular}{*{5}S}
\toprule
 \textbf{Match size} & \textbf{Processed lemmas}& \textbf{Max tokens per prompt}& \textbf{Price (\euro)} & \textbf{\cent\euro/lemma}\\
\midrule
 1 & 400 & 100 & 0.60 & 0.1500\\
 \midrule
 3 & 1179 & 500 & 0.78 & 0.0662\\
  \midrule
 5 & 1290 & 1000 & 0.84 & 0.0651\\
  \midrule
 10 & 1650 & 2000 & 0.90 & 0.0545\\
\bottomrule
\end{tabular}
\caption{Batch sizes experiments per half an hour.}
\label{tab:batch-size}
\end{table*}

The approach to generate the definitions consists of a prompt-based GPT-3 query submitted to the OpenAI API. The model selected for the generation is "text-davinci-00". Our initial prompt was: \textit{Generate in Spanish a definition of the word "[word]"}. However, we quickly realized that this could have been more optimal in terms of time and money. Before submitting the entire amount of lemmas, we tested different approaches for half an hour each, as shown in Table \ref{tab:batch-size}. We improved the performance and produced the whole dictionary in around \num{30} hours for \num{40} euros. The model parameter named temperature, also known as creativity, is set by default at \num{0.5} in all experiments. As reported, a bigger batch also implies a higher maximum of generated tokens to fit the entire content. Increasing the output length does not produce equally more extended definitions, in any case.

The first version of the "Spanish Built Factual Freectianary" (Spanish-BFF) is available at the Hugging Face\footnote{\url{https://huggingface.co/datasets/MMG/spanishBFF}} hub and at our Github\footnote{\url{https://github.com/dezzai/Spanish-BFF}}.

\section{Results\label{sec:results}}

To explore the quality of the generated dictionary, we do a qualitative analysis of definitions. We use the BLEU, Levenshtein and Jaccard scores and a sentence-transformers model \cite{reimers19} to analyze those definitions quantitatively, and eventually, we perform manual error analysis.

\subsection{Qualitative analysis}

When defining words, GPT-3 demonstrates favourable lexicographic qualities across its various versions. In the case of nouns (Appendix \ref{ex0.1}), it alludes to its class. As seen in Appendix \ref{ex0.2}, another verb is usually employed when defining them. Regarding adjectives (Appendix \ref{ex0.3}), structures like "que..." ("that...") and "se refiere a..." ("it refers to...") or synonyms of the defined word are common. Adverbs are described using "de manera..." ("in a [adjective] way") or with synonyms (Appendix \ref{ex0.4}). Neologisms are remarkably defined when they are borrowed from English terms and when they have been generated by morphological processes, as seen in Appendix \ref{ex0.5}.

\subsection{Quantitative analysis}

We divide the quantitative evaluation of the definitions into two approaches. Firstly, we use the BLEU score \cite{papineni02}, Levenshtein distance, Jaccard index, and cosine similarity for lemmas with just one meaning to inspect the quality of the description, as in \cite{noraset17}. Secondly, when further entries (polysemous words) are available, we use cosine similarity to rank the purpose with the highest resemblance to the generated one. From our lemmas, \num{44554} have one definition, and \num{21799} are polysemous in DLE. As previously stated, we query DLE to retrieve its reliable reference definitions for the sole purpose of this contrast. We use alternative handpicked definitions from other resources for neologisms that still need to be added to DLE.

\subsubsection{Monosemy}

\begin{table}[t]
\centering
\begin{tabular}{lS}
 \toprule
 \textbf{Metric} & \textbf{Score}\\
 \midrule
 {Cumulative BLEU} & 0.0083 \\
 \midrule
 {1-gram BLEU} & 0.1069 \\
 \midrule
 {Levenshtein} & 53.22 \\
 \midrule
 {Jaccard} & 0.0812 \\
 \bottomrule
\end{tabular}
\caption{Metrics for lemmas with just one sense in DLE.}
\label{tab:one-sense-metrics}
\end{table}


\begin{table}[t]
\centering
\begin{tabular}{ll*{2}S}
 \toprule
 & \textbf{Measure} & \textbf{Mean} & \textbf{Std Dev}\\  
 \midrule
 \multirow{2}[3]{*}{\textbf{DLE}} & words & 10.2 & 8.7 \\
 \cmidrule(lr){2-4}
 & characters & 59.9 & 50.2 \\
 \midrule
 \multirow{2}[3]{*}{\textbf{Spanish-BFF}} & words & 8.3 & 5.1 \\
 \cmidrule(lr){2-4}
 & characters & 49.1 & 28.4 \\
 \bottomrule
\end{tabular}
\caption{Statistical distribution (mean and standard deviation) of the definitions' lengths.}
\label{tab:definitions-length}
\end{table}

There exist \num{44554} lemmas with a single definition, with metrics reported in Table \ref{tab:one-sense-metrics}. As expected, the cumulative BLEU score (\mbox{1-grams}, \mbox{2-grams}, \mbox{3-grams}, and \mbox{4-grams} equally weighted) is deficient. As a reference,  best model achieves \num{35.78} for English \cite{noraset17}. Levenshtein's and Jaccard's scores also exhibit flawed values. Undoubtedly, the gold definitions we use to compare the generated output have a very high standard. Besides, on average, GPT-3 definitions are shorter than the corresponding DLE entries, as shown in Table \ref{tab:definitions-length}. And DLE also contains additional information, such as POS tags, etymology, or domain, that is out of this procedure's scope and should be withdrawn. In addition, GPT-3 was presumably trained on diverse texts, some of which likely have less lexicographic quality. Consequently, some lexicographical rules and errors are prominently reprised, see section \ref{sec:error-analysis}.

\begin{table}[t]
\centering
\begin{tabular}{l*{2}S}
 \toprule
 \textbf{POS tag} & \textbf{Cosine Similarity} & \textbf{\% of total}\\
 \midrule
 {All} & 0.3598 & 100 \\
 \midrule
 {Nouns} & 0.2886 & 57.41 \\
 \midrule
 {Adjectives} & 0.4746 & 26.26 \\
 \midrule
 {Verbs} & 0.3866 & 14.01 \\
 \midrule
 {Adverbs} & 0.6623 & 2.32 \\
 \bottomrule
\end{tabular}
\caption{Cosine similarity for lemmas with just one sense in DLE.}
\label{tab:cosine-similarity}
\end{table}

We use the model named "distiluse-base-multilingual-cased-v2" \footnote{\url{https://huggingface.co/sentence-transformers/distiluse-base-multilingual-cased}} to provide the cosine similarity using sentence-transformers, see Table \ref{tab:cosine-similarity}. Since this metric relies on contextual embeddings, the results positively improve. Specifically, GPT-3 performs better defining adverbs or adjectives than verbs or nouns. 

\subsubsection{Polysemy}

\begin{figure*}[t]
  \centering
  \includegraphics[width=0.95\textwidth]{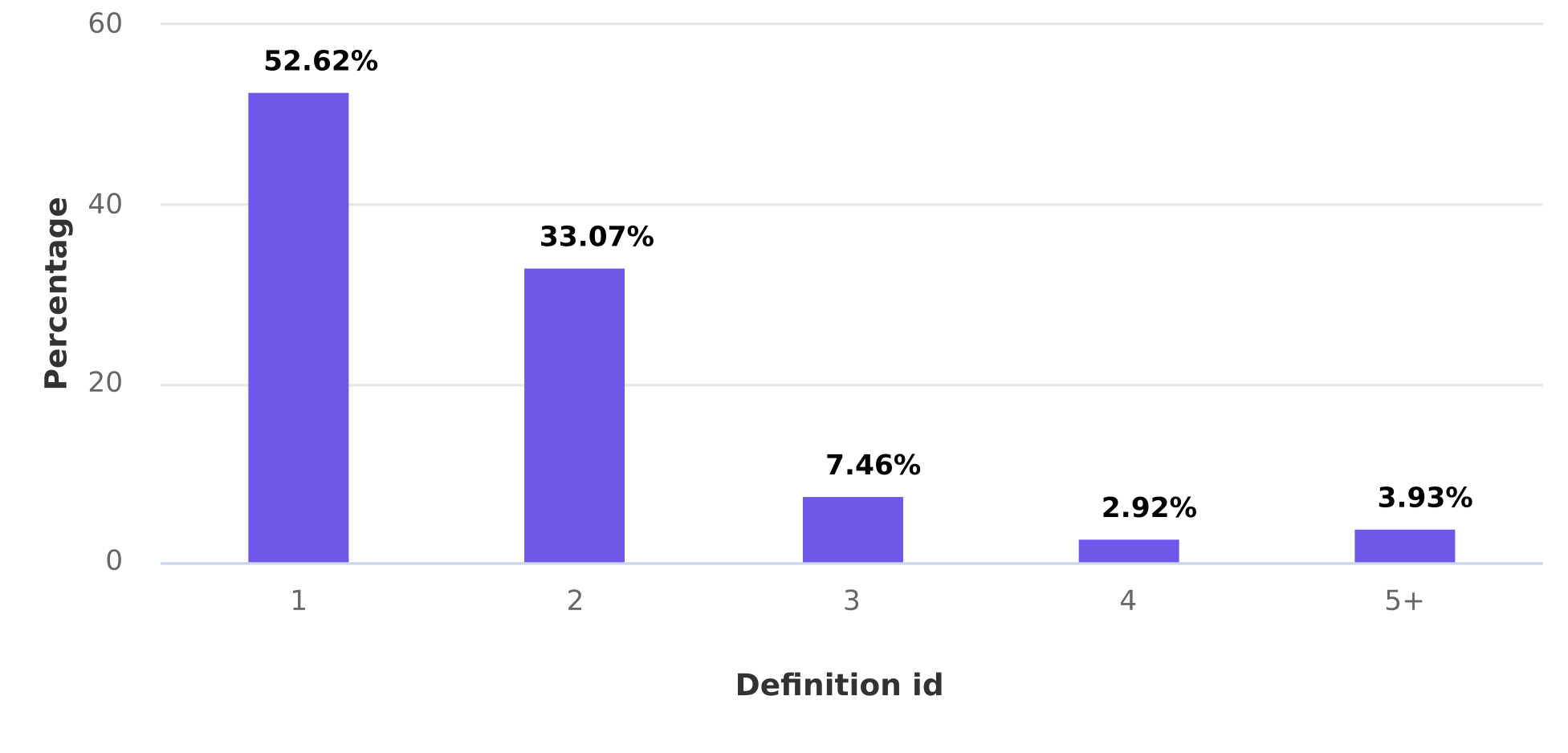}
  \caption{Highest cosine similarity among predictions and DLE definitions by ID}
  \label{fig:cosine-similarity-id}
\end{figure*}

There are \num{21799} lemmas with two or more meanings. As seen in Figure \ref{fig:cosine-similarity-id}, the likelihood of matching deeper definitions decreases rapidly. This decline makes sense since both dictionaries are based on the frequency of use. Yet, some DLE lemmas rely on the chronological order, disrupting the expected ranking. While shallow definition matches indicate good health (as GPT-3 is quicker at capturing the statistical trend of meanings), deeper matches show the tergiversation of the mainstream connotations. On average, the most similar definition, from all the senses provided for a lemma at DLE, has a mean cosine similarity of \num{0.443} with the generated one, which is higher than for monosemous words, as shown in Table \ref{tab:cosine-similarity}.

\subsection{Error analysis\label{sec:error-analysis}}

By manual error analysis, we obtained the following type of errors:

\begin{itemize}
    \item Around \num{11}~\% of the definitions start with "A [lemma] is...". When defining a word, whatever the defined lemma, it must not appear in the description (see Appendix \ref{ex1.1}). But, this straightforward formula does not involve circular logic. Thus, it is easy to post-process these cases to deliver better results.
    \item Sometimes, a word is defined as an almost similar spelt word: for instance, "re", which could be described as a musical note, could also be defined as "monarca" instead, a synonym of "rey", "king" in Spanish. In Appendix \ref{ex1.2}, we can inspect some examples of the drawbacks of subword tokenizing, which are consolidated using quotation marks around the lemmas to avoid splitting.
    \item When nouns are defined as conjugated verbs (Appendix \ref{ex1.3}), for instance, "pare" ("stop") is a noun derived from the verb "parar" ("to stop"), but GPT-3 defines it as the verb. 
    \item There are cases of language interference (Appendix \ref{ex1.4}) in which the Spanish word is poorly defined in English, primarily due to highly uncommon lemmas.
    \item Occasionally, part of the meaning is accurate (Appendix \ref{ex1.5}). For instance, "maquis", a Spanish guerrilla during Civil War, is defined by GPT-3 as where this guerrilla used to fight. 
    \item Despite all, some terms are just completely wrong. For example, the word "yangüés" is a demonym wrongly defined as a bird (Appendix \ref{ex1.6}). 
    \item The specific meaning of some neologisms is not adequately captured, and the definition alludes to a related term instead (Appendix \ref{ex1.7}). 

\end{itemize}

Indeed, most of these errors can be partially solved with better prompting. Queries such as \textit{Genera en español la definición para la palabra literal "[palabra]"} (\textit{Generate in Spanish the literal definition for the word "[word]"}) averts the subword tokenizing and skips some nouns-as-verbs errors. 

\section{Future\label{sec:future-lines}}

We understand this is a naive approach to creating a dictionary and must fix numerous lexicographic rules. However, this was just the first attempt to commit to the task, which we consider extremely useful for NLP and lexicography. In the future, we aim to solve these issues by, for instance, taking homonymy and polysemy into account, using POS tags, semantics roles, example sentences, domain, and more. We also plan to do the same for other languages. Furthermore, we intend to use ChatGPT and other instruct models to find the best approach.

\section{Conclusions\label{sec:conclusions}}

Building a dictionary is a complex task that, to our knowledge, has yet to be fully explored with LLMs. Here we introduce the first Spanish dictionary generated with GPT-3. The "freectianary" mourns some flaws (some lexicographic rules are broken and other mistakes related to language models) but also contains some promising aspects (reliance on some different lexicographic rules). In this paper, we define the future steps we aim to follow to improve our initial commitment to the task, which goes from improving the Spanish dictionary to generating dictionaries for other languages. 

\section{Limitations\label{sec:limitations}}

This approach is based on a list of lemmas, so we understand it is limited to languages with substantial resources, such as Spanish. Another option could be using a corpus and a lemmatizer, but we should note that not all languages have these resources.

\section{Ethics statement\label{sec:ethics}}

We understand the possibilities that models like GPT-3 can imply for industry and future academic research. We intend to contribute to a better understanding and development of NLP and promote responsible use.

\bibliography{anthology,custom}

\begin{thebibliography}{15}
\expandafter\ifx\csname natexlab\endcsname\relax\def\natexlab#1{#1}\fi

\bibitem[{Bergenholtz and Gouws(2012)}]{bergenholtz12}
Henning Bergenholtz and Rufus~H. Gouws. 2012.
\newblock What is lexicography?
\newblock \emph{Lexikos}, 22:31--42.
\newblock ISBN: 2224-0039.

\bibitem[{Bevilacqua et~al.(2020)Bevilacqua, Maru, and Navigli}]{bevilacqua20}
Michele Bevilacqua, Marco Maru, and Roberto Navigli. 2020.
\newblock Generationary or “how we went beyond word sense inventories and
  learned to gloss”.
\newblock In \emph{Proceedings of the 2020 Conference on Empirical Methods in
  Natural Language Processing (EMNLP)}, pages 7207--7221.

\bibitem[{Brown et~al.(2020)Brown, Mann, Ryder, Subbiah, Kaplan, Dhariwal,
  Neelakantan, Shyam, Sastry, and Askell}]{brown20}
Tom Brown, Benjamin Mann, Nick Ryder, Melanie Subbiah, Jared~D. Kaplan,
  Prafulla Dhariwal, Arvind Neelakantan, Pranav Shyam, Girish Sastry, and
  Amanda Askell. 2020.
\newblock Language models are few-shot learners.
\newblock \emph{Advances in neural information processing systems},
  33:1877--1901.

\bibitem[{Hausmann and Wiegand(1989)}]{hausmann89}
Franz~Josef Hausmann and Herbert~Ernst Wiegand. 1989.
\newblock Component parts and structures of general monolingual dictionaries: A
  survey.
\newblock \emph{1989-1991}.

\bibitem[{Hill et~al.(2016)Hill, Cho, Korhonen, and Bengio}]{hill16}
Felix Hill, Kyunghyun Cho, Anna Korhonen, and Yoshua Bengio. 2016.
\newblock Learning to understand phrases by embedding the dictionary.
\newblock \emph{Transactions of the Association for Computational Linguistics},
  4:17--30.
\newblock Publisher: MIT Press.

\bibitem[{Jackson(2013)}]{jackson13}
Howard Jackson. 2013.
\newblock \emph{Lexicography: an introduction}.
\newblock Routledge.

\bibitem[{Kirkness(2004)}]{kirkness04}
Alan Kirkness. 2004.
\newblock Lexicography.
\newblock \emph{The handbook of applied linguistics}, pages 54--81.

\bibitem[{Lewis et~al.(2019)Lewis, Liu, Goyal, Ghazvininejad, Mohamed, Levy,
  Stoyanov, and Zettlemoyer}]{lewis19}
Mike Lewis, Yinhan Liu, Naman Goyal, Marjan Ghazvininejad, Abdelrahman Mohamed,
  Omer Levy, Ves Stoyanov, and Luke Zettlemoyer. 2019.
\newblock Bart: Denoising sequence-to-sequence pre-training for natural
  language generation, translation, and comprehension.
\newblock \emph{arXiv preprint arXiv:1910.13461}.

\bibitem[{Malkin et~al.(2021)Malkin, Lanka, Goel, Rao, and Jojic}]{malkin21}
Nikolay Malkin, Sameera Lanka, Pranav Goel, Sudha Rao, and Nebojsa Jojic. 2021.
\newblock {GPT} {Perdetry} {Test}: {Generating} new meanings for new words.
\newblock In \emph{Proceedings of the 2021 {Conference} of the {North}
  {American} {Chapter} of the {Association} for {Computational} {Linguistics}:
  {Human} {Language} {Technologies}}, pages 5542--5553.

\bibitem[{Noraset et~al.(2017)Noraset, Liang, Birnbaum, and Downey}]{noraset17}
Thanapon Noraset, Chen Liang, Larry Birnbaum, and Doug Downey. 2017.
\newblock Definition modeling: {Learning} to define word embeddings in natural
  language.
\newblock In \emph{Proceedings of the {AAAI} {Conference} on {Artificial}
  {Intelligence}}, volume~31.
\newblock Issue: 1.

\bibitem[{Ortega-Mart{\'\i}n(2021)}]{ortega21}
Miguel Ortega-Mart{\'\i}n. 2021.
\newblock \emph{Grafos de vinculaci{\'o}n sem{\'a}ntica a partir del definiens
  del DUE}.
\newblock Ph.D. thesis, Universidad Complutense de Madrid.

\bibitem[{Ouyang et~al.(2022)Ouyang, Wu, Jiang, Almeida, Wainwright, Mishkin,
  Zhang, Agarwal, Slama, and Ray}]{ouyang22}
Long Ouyang, Jeff Wu, Xu~Jiang, Diogo Almeida, Carroll~L. Wainwright, Pamela
  Mishkin, Chong Zhang, Sandhini Agarwal, Katarina Slama, and Alex Ray. 2022.
\newblock Training language models to follow instructions with human feedback.
\newblock \emph{arXiv preprint arXiv:2203.02155}.

\bibitem[{Papineni et~al.(2002)Papineni, Roukos, Ward, and Zhu}]{papineni02}
Kishore Papineni, Salim Roukos, Todd Ward, and Wei-Jing Zhu. 2002.
\newblock Bleu: a method for automatic evaluation of machine translation.
\newblock In \emph{Proceedings of the 40th annual meeting of the Association
  for Computational Linguistics}, pages 311--318.

\bibitem[{Reimers and Gurevych(2019)}]{reimers19}
Nils Reimers and Iryna Gurevych. 2019.
\newblock Sentence-bert: Sentence embeddings using siamese bert-networks.
\newblock \emph{arXiv preprint arXiv:1908.10084}.

\bibitem[{Schmidt(2019)}]{schmidt19}
Robin~M Schmidt. 2019.
\newblock Recurrent neural networks (rnns): A gentle introduction and overview.
\newblock \emph{arXiv preprint arXiv:1912.05911}.

\end{thebibliography}
\bibliographystyle{acl_natbib} 

\appendixtitleon
\appendixtitletocon
\begin{appendices}

\section{Adequate examples\label{sec:examples}}

\subsection{Nouns} 
\label{ex0.1}
\begin{itemize}
    \item "exotismo": "La cualidad de lo que es exótico o extraño, especialmente en relación con la cultura y la naturaleza." ("exoticism": "The quality of what is exotic or strange, especially in relation to culture and nature.")
    \item  "camisón": "Una prenda de vestir ligera y suelta, normalmente de algodón, que se usa para dormir." ("nightgown": "A light, loose-fitting article of clothing, usually cotton, worn for sleeping.")
    \item "pillería": "Acto de robar o tomar algo sin permiso." ("mischief": "Act of stealing or taking something without permission.")
\end{itemize}

\subsection{Verbs}
\label{ex0.2}
\begin{itemize}
    \item "guiar": "dirigir una persona o un grupo." ("to lead": "direct a person or a group.")
    \item "parir": "dar a luz a un bebé." ("to give birth": "to give birth to a baby."
    \item "coagular": "formar coágulos." ("to coagulate": "to form clots.")
\end{itemize}

\subsection{Adjectives}
\label{ex0.3}
\begin{itemize}
    \item "arrendable": "Que se puede alquilar." ("rentable": "That can be rented.")
    \item "candente": "Ardiente o intenso." ("glowing": "Fiery or intense.") 
\end{itemize}

\subsection{Adverbs}
\label{ex0.4}
\begin{itemize}
    \item "peculiarmente": "de manera particular." ("peculiarly": "in a particular way.")
    \item "otrora": "antiguamente." ("once": "formerly.")
\end{itemize}

\subsection{Neologisms}
\label{ex0.5}
\begin{itemize}
    \item "antibotellón": "Antibotellón es una iniciativa para prevenir la reunión de grandes grupos de personas en espacios públicos para beber alcohol y evitar los disturbios asociados." (Antibotellón is an initiative to prevent the gathering of large groups of people in public spaces to drink alcohol and avoid the associated riots)
    \item  "mainstream": "Lo que es ampliamente aceptado o popular en una cultura, una comunidad o una sociedad." ("mainstream": "What is widely accepted or popular in a culture, a community or a society." )
    \item "phising": "El phising es una forma de fraude cibernético que consiste en el uso de correos electrónicos o mensajes de texto fraudulentos para obtener información confidencial, como nombres de usuario y contraseñas." ("phising": "Phishing is a form of cyber fraud that involves the use of fraudulent email or text messages to obtain sensitive information such as usernames and passwords.")
    \item "fomo": "FOMO es un acrónimo de "miedo a perderse", y se refiere a la ansiedad que uno puede sentir cuando ve que otros están disfrutando de algo que él o ella no está haciendo." ("fomo": "FOMO is an acronym for "fear of missing out", and refers to the anxiety one may feel when seeing others enjoying something that he or she is not doing.")
    \item "nomofobia": "Es el miedo excesivo a estar desconectado de la tecnología, especialmente de los teléfonos inteligentes." ("nomophobia": "It is the excessive fear of being disconnected from technology, especially smartphones.")
    \item "googlear": "Buscar información en Internet usando el motor de búsqueda de Google." ("to google": "Search for information on the Internet using the Google search engine.")
    \item "polimedicar": "Polimedicar es un término médico que se refiere a la práctica de tratar a un paciente con varias medicinas o terapias al mismo tiempo." ("polymedicate": "Polymedicate is a medical term that refers to the practice of treating a patient with several medicines or therapies at the same time.")
\end{itemize}

\section{Errors\label{sec:errors}}

\subsection{Lack of lexicographic quality}
\label{ex1.1}
\begin{itemize} 
    \item "campeonato": "Un campeonato es una competición para determinar quién es el mejor en algo." ("championship": "A championship is a competition to determine who is the best at something.")
    \item "negacionismo": "El negacionismo es una actitud consistente en negar o discutir la existencia de un hecho o una realidad." ("denialism": "Denialism is an attitude consisting of denying or discussing the existence of a fact or reality.")
\end{itemize}

\subsection{Tokenizer issues}
\label{ex1.2}
\begin{itemize}
    \item "re": "monarca" ("re": "monarch")
    \item "tona": "Una tona es una medida de volumen equivalente a 1000 kilogramos." ("cream": "A ton is a measure of volume equal to 1000 kilograms.")
    \item "adoquiera": "adquirir algo." ("anywhere": "to acquire something.")
\end{itemize}

\subsection{Noun as a conjugated verb}
\label{ex1.3}
\begin{itemize}
    \item "paragranizo": "Verbo que significa proteger con una granizada." (*hail net: "Verb that means to protect with a hailstorm.")
    \item "pare": "Pare es un verbo que significa detener algo, detener a alguien o interrumpir una acción." ("stop -noun-": "Stop is a verb that means to stop something, stop someone, or interrupt an action.")
    \item "lleve": "El verbo llevar significa transportar algo de un lugar a otro." (*money from public transactions: "The verb to carry means to transport something from one place to another.")
\end{itemize}

\subsection{English used in the definition}
\label{ex1.4}
\begin{itemize}
    \item "relva": "Grass." (*Raise the fallow.: "Grass")
    \item "sueldacostilla": "Paycheck." (*A plant: "Paycheck")
\end{itemize}

\subsection{Somehow related}
\label{ex1.5}
\begin{itemize}
    \item "maquis": "Una zona de bosque o maleza en la que los combatientes se ocultan para luchar contra una ocupación militar." (*Spanish guerrillas: "An area of forest or brush where combatants hide to fight against a military occupation.")
    \item "baleario": "Una región insular del mar Mediterráneo, compuesta por las islas Baleares." ("balearic": "An insular region of the Mediterranean Sea, made up of the Balearic Islands.")
    \item "zampabollos": "Persona que se dedica a vender bollos por la calle." (*someone who eats too much: "Person who is dedicated to selling buns on the street.")

\end{itemize}

\subsection{Just bad}
\label{ex1.6}
\begin{itemize}
    \item "ll": "Abreviatura de la expresión \"Llámame\"." ("ll": "Abbreviation for the expression \"Call me\".")
    \item "yangüés": "Pájaro de la familia de los estorninos." (*a demonym: "Bird of the starling family.")
    \item "napoleón": "Napoleon."
    \item "menos": "menos." ("less": "less.")
\end{itemize}

\subsection{Neologisms}
\label{ex1.7}
\begin{itemize}
    \item "pagafantas": "Persona que invita a los demás a salir o a tomar algo sin tener intención de pagar la cuenta." ("simp": "Person who invites others to go out or have a drink without intending to pay the bill.")
    \item "telonear": "Ver una película en una pantalla grande, como una sala de cine." ("playing before the main band in a concert": "Watch a movie on a big screen, like a movie theater.")
    \item "chandalismo": "El chandalismo se refiere al comportamiento de personas que, sin ningún motivo, destruyen o dañan bienes públicos o privados." ("use and abuse of wearing tracksuit": "Chandalism refers to the behavior of people who, without any reason, destroy or damage public or private property." )
\end{itemize}

\end{appendices}

\end{document}